\documentclass[10pt,twocolumn,letterpaper]{article}

\usepackage[pagenumbers]{cvpr} 

\usepackage{duckuments}

\definecolor{cvprblue}{rgb}{0.21,0.49,0.74}
\usepackage[pagebackref,breaklinks,colorlinks,allcolors=cvprblue]{hyperref}

\def\paperID{6111} 
\def\confName{CVPR}
\def\confYear{2026}

\title{VLIC: Vision-Language Models As Perceptual Judges\\ for Human-Aligned Image Compression}

\newcommand{\modelname}{VLIC}

\author{
Kyle Sargent$^{1,2}$,~
Ruiqi Gao$^3$,~
Philipp Henzler$^2$,~
Charles Herrmann$^3$,~
Aleksander Hoły\'nski$^3$~\\
Li Fei-Fei$^1$,~
Jiajun Wu$^1$,~
Jason Zhang$^2$
\\[0.5em]
\textsuperscript{1}Stanford University,~
\textsuperscript{2}Google Research,~
\textsuperscript{3}Google DeepMind
}

\begin{document}

\twocolumn[{%
\maketitle
\thispagestyle{empty}
\begin{center}
\centering
\captionsetup{type=figure}
\includegraphics[width=\textwidth]{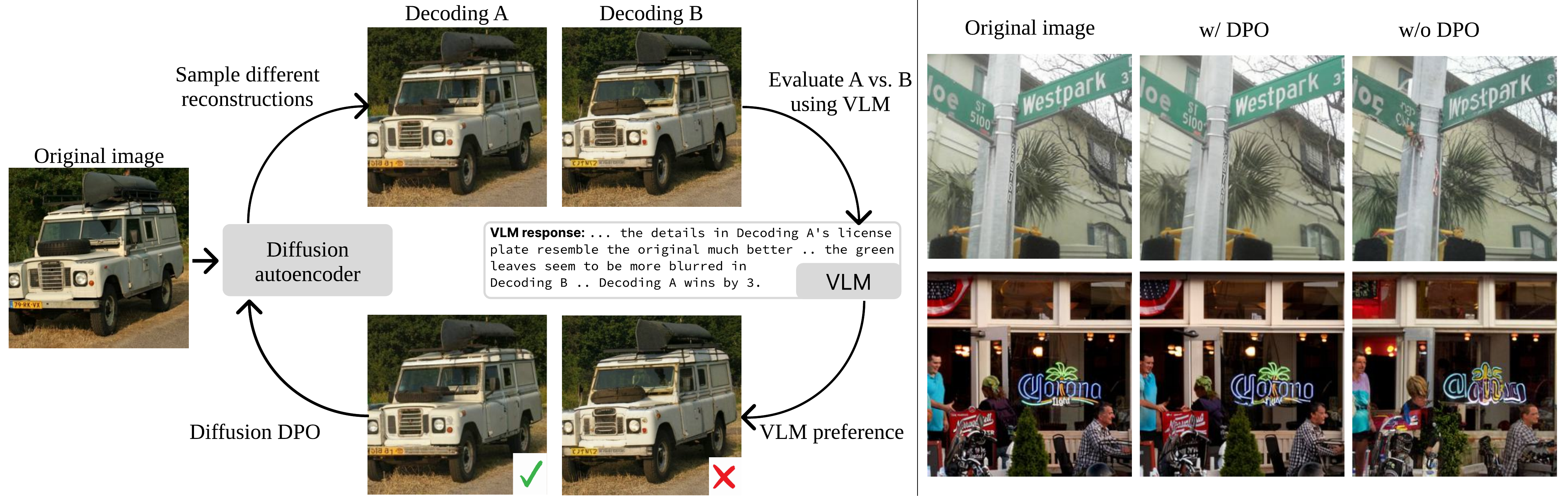}
\captionof{figure}{ \textbf{Left.} We propose a new post-training technique for diffusion  autoencoders which uses Vision-Language Models to judge different decodings of the same image and leverage these judgements to improve the autoencoder through Diffusion DPO. \textbf{Right.} Our method, VLIC, demonstrates substantial improvements in the overall reconstruction quality, as well as better alignment to human perception.
}%
\label{fig:teaser-fig}%
\end{center}%
}]

\begin{abstract}
Evaluations of image compression performance which include human preferences have generally found that naive distortion functions such as MSE are insufficiently aligned to human perception.
In order to align compression models to human perception, prior work has employed differentiable perceptual losses consisting of neural networks calibrated on large-scale datasets of human psycho-visual judgments. We show that, surprisingly, state-of-the-art vision-language models (VLMs) can replicate binary human two-alternative forced choice (2AFC) judgments zero-shot when asked to reason about the differences between pairs of images. Motivated to exploit the powerful zero-shot visual reasoning capabilities of VLMs, we propose 
\textbf{V}ision-\textbf{L}anguage Models for \textbf{I}mage \textbf{C}ompression (VLIC), a diffusion-based image compression system designed to be post-trained with binary VLM judgments. \modelname~leverages existing techniques for diffusion model post-training with preferences, rather than distilling the VLM judgments into a separate perceptual loss network. We show that calibrating this system on VLM judgments produces competitive or state-of-the-art performance on human-aligned visual compression depending on the dataset, according to perceptual metrics and large-scale user studies. We additionally conduct an extensive analysis of the VLM-based reward design and training procedure and share important insights. More visuals are available on our \href{https://kylesargent.github.io/vlic}{website}.

\end{abstract}    
\vspace{-6mm}
\section{Introduction}
\vspace{-2mm}
\label{sec:intro}


\begin{figure*}
    \centering
    \includegraphics[width=.89\linewidth]{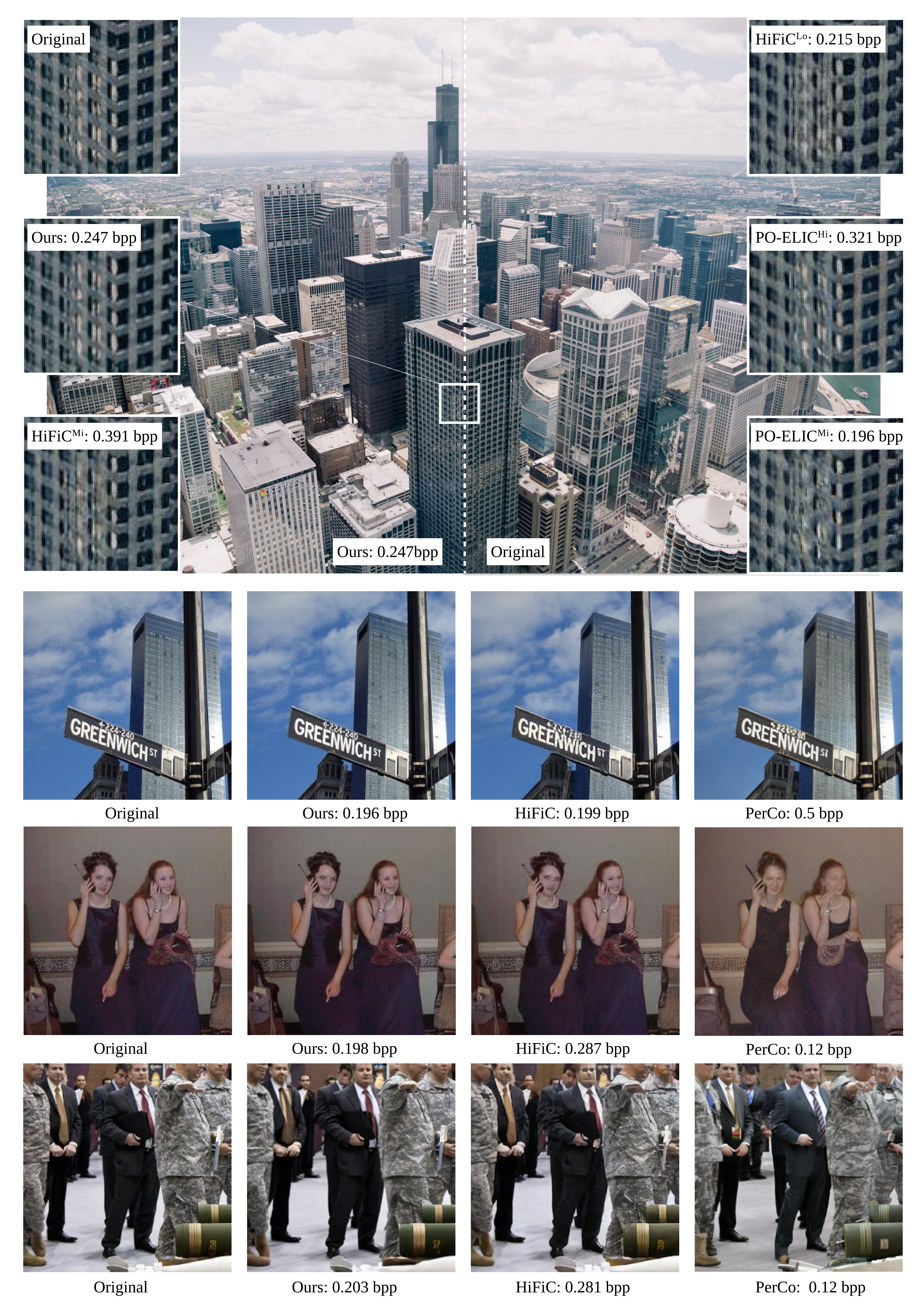}
    \vspace{-4mm}
    \caption{\textbf{Qualitative results on standard image compression datasets.} \textit{Top}: We compare \modelname~ with HiFiC \cite{mentzer2020high} and PO-ELIC \cite{he2022poelic} on a CLIC 2022 image \cite{CLIC2022Dataset} at various bits per pixel (bpp). 
    \textit{Bottom}: We compare our approach with HiFiC and PerCo on MS-COCO~\cite{lin2014microsoft}. We find that our approach represents perceptually relevant fine details, faces, and textures more faithfully}
    \label{fig:qualitative_results}
\end{figure*}

Compressing images and videos is necessary for storing and transmitting the rich sensory multimedia captured every day. This process inherently involves making trade-offs between compression rate (\ie file size) and visual quality. Ideally, the assessment of visual quality should align well with human perception, prioritizing details to which humans are more sensitive. For instance, humans are sensitive to perturbations to faces and text, but not to high-entropy natural textures such as grass or fur.

Historically, visual quality has been assessed via classical metrics such as PSNR and SSIM~\cite{wang2004ssim}. However, these metrics are poorly aligned with human perception and often contradict human judgments of visual quality~\cite{zhang2018unreasonable,ding2020dists}. To address this, prior work has focused on learning neural network approximations to human judgment of visual similarity, with varying approaches that target low-level similarity \cite{zhang2018unreasonable, ding2020dists} and semantic or high-level similarity \cite{fu2023dreamsim}. 
While these learned perceptual metrics enable training neural compression methods that are better aligned with human perception, they are not without issues.
Directly optimizing on these perceptual metrics can exploit their null-space, improving network performance with only limited gains in actual human-perceived quality~\cite{kettunen2019elpips, chen2025dito}. Moreover, trained perceptual metrics do not necessarily generalize beyond the datasets of human judgments used to calibrate them. For instance, networks calibrated on low-level visual differences may not agree with human judgments on images with high-level semantic 
differences~\cite{fu2023dreamsim}.



In this work, we propose an alternative to the dominant paradigm of learned and differentiable  perceptual metrics used during image compression network training. First, we show that Vision-Language Models (VLMs) are effective zero-shot perceptual judges of visual similarity (Figure~\ref{fig:teaser-fig}). 
Specifically, we show that an off-the-shelf VLM (Gemini 2.5-Flash~\cite{comanici2025gemini}) can replicate human judgments on multiple human visual judgment datasets, namely on BAPPS \cite{kettunen2019elpips} and our own dataset of human judgments collected on images compressed by various compression baselines.


The fact that VLM reasoning can replicate human perceptual judgments is an encouraging finding because, as VLMs are improved through considerable investments, improved automatic perceptual judges may result without additional effort to collect human judgment data and train perceptual metrics. However, it is not clear how to convert the binary 2AFC judgments produced by VLMs into an optimizable perceptual metric which can be exploited by existing GAN-based perceptually oriented compression systems.

Therefore, motivated by a desire to maximally exploit VLMs for human-oriented visual compression, we instead design a diffusion-based visual compression system similar to recent diffusion-based approaches for visual compression~\cite{sargent2025flowmodemodeseekingdiffusion, hoogeboom2023hfd, careil2023perco}. Since diffusion-based visual techniques can leverage the rich existing literature on diffusion model post-training with preferences \cite{wallace2023diffusion}, we can benefit from VLM perceptual judgments without having to use them to train a separate perceptual metric.

Concretely, we make the following contributions:
\begin{enumerate}
    \item We show that an off-the-shelf VLM (Gemini 2.5-Flash~\cite{comanici2025gemini}) can replicate human judgments of visual similarity on multiple human judgment datasets zero-shot.
    \item We present a diffusion-based visual compression system based on FlowMo \cite{sargent2025flowmodemodeseekingdiffusion} extended with an additional entropy coder. We show that VLM-generated preferences, can be used to post-train this system via Diffusion DPO \cite{wallace2023diffusion}, improving performance. Moreover, we show that ensembling VLM preferences with those of a traditional perceptual metric, LPIPS, adds additional benefits and exceeds the performance of post-training with either reward alone.
    \item We quantitatively study \modelname~and show it achieves either competitive or state-of-the art performance relative to strong existing compression baselines, depending on the dataset, and conduct several large-scale user studies. We additionally provide several empirical analyses outlining best practices for VLM-based compression post-training, conduct ablation studies on reward design, and analyze and discuss important failure modes.
\end{enumerate}

\section{Related Work}
\label{Related Work}
\vspace{-7mm}

\noindent\paragraph{Perceptually-oriented Image Compression.} Prior work has considered GAN-based~\cite{heusel2017gans} and diffusion-based~\cite{ho2020denoising} perceptually-oriented image compression. GAN-based techniques such as HiFiC \cite{mentzer2020high} and PO-ELIC \cite{he2022poelic} are quite popular and are often fast to decode, particularly if a factorized entropy model is used \cite{minnen2020channel, he2022poelic}. Diffusion-based visual compression can be separated into two distinct styles. The first style, based on diffusion autoencoders \cite{preechakul2021diffusion}, attempts to learn perceptually-oriented visual compression end-to-end with a discrete latent bottleneck \cite{xu2024disco, birodkar2024sample, sargent2025flowmodemodeseekingdiffusion, chen2025dito, bachmann2025flextok, yang2023mandt} or with the decoder conditioned on a prior learnt representation \cite{hoogeboom2023hfd}. This style sometimes has the added benefit of producing a tractable latent space for downstream generative modeling. The second style involves using a trained diffusion model as an entropy coder over the diffusion model's reverse process \cite{ohayon2025ddcm, theis2022diffc}. Our model architecture belongs to the first style of diffusion-based visual compression, and is derived from FlowMo \cite{sargent2025flowmodemodeseekingdiffusion} with a few modifications, while our training scheme is a novel combination of diffusion autoencoder training with Diffusion DPO~\cite{black2023training}.

\vspace{-4mm}
\noindent\paragraph{Approximations to Human Perception of Visual Similarity.}
Various work has explored designing proxies for human perception, such as LPIPS~\cite{zhang2018unreasonable}, E-LPIPS~\cite{kettunen2019elpips}, DreamSim~\cite{fu2023dreamsim}, and DISTS~\cite{ding2020dists}. These models have been used to calibrate learned image compression models \cite{mentzer2020high, he2022poelic} but have also been used to generate different human visual content which humans will perceive similarly \cite{freeman2011metamers}. In visual compression, other work has argued for the use of text in the encoding process together with a generative decoder \cite{weissman2023toward}, with the intuition that text better captures human-relevant information in visual data. Moreover, practical implementations of text-aligned visual compression have been realized \cite{lu2025atoken, lee2024taco, lei2023textsketch}, though techniques leveraging self-supervised learning backbones \cite{yao2025lightningdit} and generative foundation models \cite{careil2023perco} have also found success. Different from these works, we directly train a compression system using a VLM as a zero-shot proxy for human judgment of visual similarity, eschewing learned metrics calibrated on human preferences \cite{zhang2018unreasonable, fu2023dreamsim} or heuristics for which data to encode \cite{lee2024taco, lei2023textsketch}.

\vspace{-4mm}
\noindent\paragraph{Aligning Diffusion Models to Preferences.} Various works have considered how to align diffusion models to preferences. Differentiable techniques include DRAFT~\cite{clark2023draft} and VADER~\cite{prabhudesai2024video}, while reinforcement learning-based techniques, such as Diffusion Direct Preference Optimization (DDPO) \cite{wallace2023diffusion} and Denoising Diffusion Policy Optimization \cite{black2023training}, can support non-differentiable preferences as rewards.

Several works have explored aligning diffusion models to rewards produced by VLMs, including RewardDance~\cite{wu2025rewarddance} and HSPv3~\cite{ma2025hpsv3}. Our choice to use VLMs to produce rewards for image compression is contextually novel in the context of image compression aligned to human preferences, and is motivated by our finding that VLMs can replicate human visual similarity judgments zero-shot, but is inspired by prior works in diffusion model post-training.


\section{Method}
\begin{figure*}
    \centering
    \includegraphics[width=\linewidth]{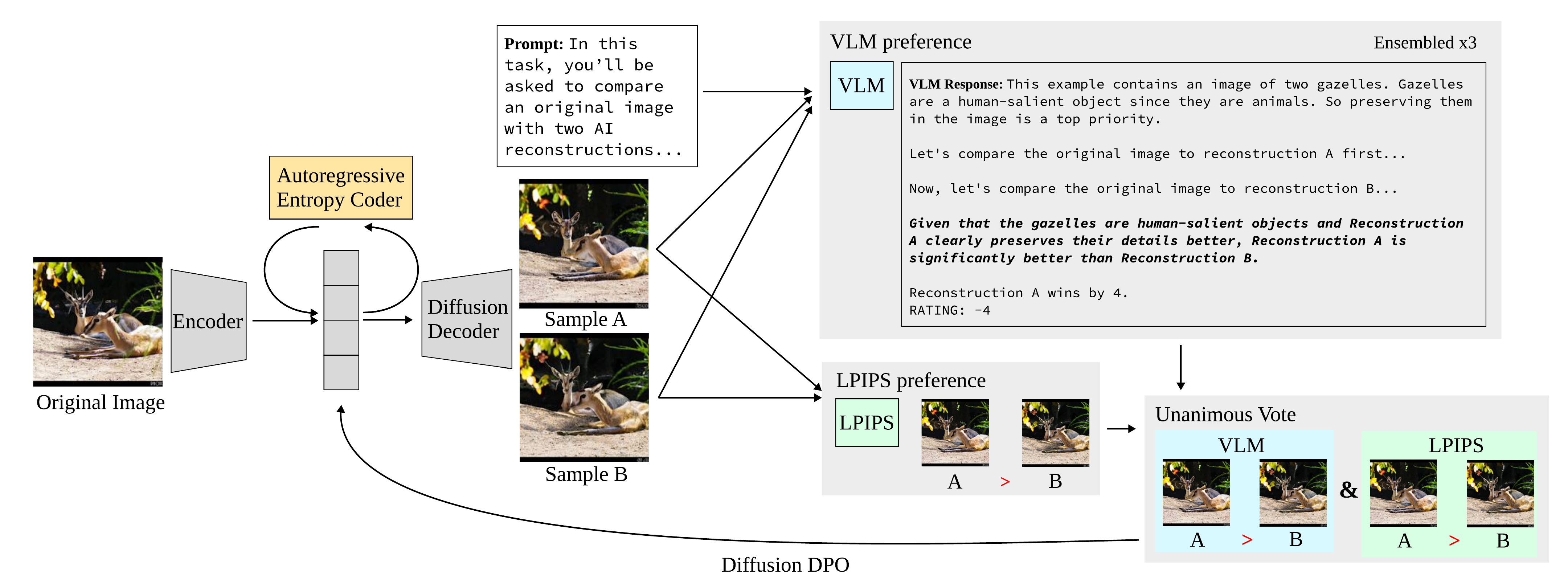}
    \caption{\textbf{Method.} An original image is encoded to a one-dimensional discrete latent code via an encoder. The discrete code is entropy coded by an auto-regressive language model. The diffusion decoder samples two reconstructions conditioned on the latent code, which are ranked via a VLM. The resulting preference is used to train the full diffusion autoencoder via Diffusion DPO~\cite{wallace2023diffusion}.}
    \label{fig:method}
    \vspace{-4mm}
\end{figure*}

We will now provide an overview of the \modelname~system. We will first review the architecture and training scheme. Then we will explain the process by which a VLM is guided to produce perceptual judgments. A full method diagram is shown in Figure \ref{fig:method}.

\noindent\paragraph{Architecture and training.} Our architecture and first training stage are identical to FlowMo \cite{sargent2025flowmodemodeseekingdiffusion}, with the only architectural difference being that we use finite scalar quantization (FSQ) \cite{Mentzer2023FSQ} in lieu of lookup-free quantization \cite{Yu2023LFQ} for simplicity and to eliminate the commitment and entropy losses. We similarly adopt the rectified flow framework \cite{liu2022flow, Lipman2022Flow} and use an LPIPS loss on the 1-step denoised prediction of the diffusion decoder following prior work \cite{yang2023mandt, sargent2025flowmodemodeseekingdiffusion, chen2025dito}. 

Different from prior work, for our second stage of training we elect to post-train the model via Diffusion DPO~\cite{wallace2023diffusion} to align the diffusion model to arbitrary (potentially non-differentiable) preferences. Diffusion DPO is a variant of Direct Preference Optimization (DPO) \cite{Rafailov2023dpo} adapted to the unique setting of diffusion. 
The Diffusion DPO objective, adapted to the discrete diffusion autoencoder setting where the denoising network encodes and quantizes the original image $\mathbf{x}$ internally, is
\begin{equation}
    L_{\textrm{DDPO}}(\theta) = -\mathbb{E} \log \sigma \left( -\beta \omega(\lambda_t)( \Delta^w - \Delta^l) \right),
\end{equation}
where
\begin{equation}
    \small{\Delta^w = \left \lVert \boldsymbol{\epsilon}^w - \boldsymbol{\epsilon}_\theta(\mathbf{\hat x}_t^w, \mathbf{x}, t)\right\rVert_2^2 
- \left \lVert\boldsymbol{\epsilon}^w - \boldsymbol{\epsilon}_{\text{ref}}(\mathbf{\hat x}_t^w, \mathbf{x}, t)\right \rVert_2^2,}
\end{equation}
\begin{equation}
    \small{\Delta^l = \left \lVert \boldsymbol{\epsilon}^l - \boldsymbol{\epsilon}_\theta(\mathbf{\hat x}_t^l, \mathbf{x}, t)\right\rVert_2^2 
- \left \lVert\boldsymbol{\epsilon}^l - \boldsymbol{\epsilon}_{\text{ref}}(\mathbf{\hat x}_t^l, \mathbf{x}, t)\right \rVert_2^2.}
\end{equation}
The expectation is taken over sampled reconstructions from the model and ranked as winner $\mathbf{\hat x_0}^w$ and loser $\mathbf{\hat x_0}^l$, and with $\beta$ a KL-weight controlling the degree to which the learned policy can deviate from the original reference policy. $\epsilon$ and $\epsilon_\theta$ are the noise and noise estimator respectively, $t$ the timestep, and $\omega(\lambda_t)$ an SNR-dependent weighting factor. $\Delta^w$ represents the difference between the loss on the winning example between the current and reference policy, which is intended to be decreased, while the loss difference on the losing example $\Delta^l$ is increased.


Unlike techniques requiring differentiable rewards~\cite{clark2023draft, prabhudesai2024video}, this formulation of Diffusion DPO can be used with our VLM-defined rewards for image quality assessment. Moreover, unlike methods such as Denoising Diffusion Policy Optimization~\cite{black2023training} which require a carefully tuned value function or baseline, Diffusion DPO trains stably across diverse datasets as winning samples are contrasted against losing samples with the same latent code as conditioning, which is similar to GRPO~\cite{shao2024deepseekmath} with $n=2$. We additionally co-train with the original flow matching training loss, since we find this allows us to post-train for longer without divergence. This loss is:
\begin{equation}
    L_{\mathrm{Flow}}(\theta) = \mathbb{E}_{\epsilon,x,t} (\|\mathbf{v} - \mathbf{v}_{\theta}(\mathbf{x}, \mathbf{x}_t, t) \|_2^2),
\end{equation}
with $\boldsymbol{v} = \boldsymbol{\epsilon} + \boldsymbol{x}$ 
the flow matching velocity and $\mathbf{v}_\theta(\mathbf{x_t}, \mathbf{x}, t)$ the velocity estimate of the diffusion autoencoder (note that we may predict either $\boldsymbol{\epsilon}$ or $\boldsymbol{v}$ via reparameterizing the network output, which is by default in $\boldsymbol{v}$-parameterization~\cite{salimans2022progressive} in our case) given the noisy original image $\mathbf{x_t}$, timestep $t$ and original image $\mathbf{x}$ which is quantized within the network. Our final training loss is:
\begin{equation}
    L(\theta) = L_{\mathrm{DDPO}}(\theta) + \lambda_{\mathrm{Flow}}L_{\mathrm{Flow}}(\theta)
\end{equation}
with hyperparameter $\lambda_{\mathrm{Flow}}$. We train with the encoder unfrozen which led to slightly better performance and may enable the encoder to acquire features necessary to improve the reward in $L_{\mathrm{DDPO}}(\theta)$, since the VLM reward is unseen during pretraining.


We further compress the discrete tokens from FSQ via a secondary entropy coder, which is trained separately. The entropy coder takes the form of a simple autoregressive transformer over the 1-dimensional latent sequence. After this entropy coder is trained, we use it to compress the latent code via arithmetic coding, similar to prior work \cite{deletang2024language}.

\noindent\paragraph{The \modelname~reward function.} \modelname~ is a diffusion autoencoder, meaning that a given image is compressed deterministically to a discrete latent code $c$, but then decompressed stochastically. Post-training with Diffusion DPO involves sampling two decompressions of the same latent code and ranking them via a reward function to produce a winning and losing sample $\boldsymbol{\hat x}_0^w$ and $\boldsymbol{\hat x}_0^l$. 

Any reward function can in principle be used to determine the winning and losing sample. In our work, we use an off-the-shelf VLM (Gemini 2.5-Flash~\cite{comanici2025gemini}) to judge decompressed images. 
An overview of the reward computation is shown on the right hand side  of Figure \ref{fig:method}. Essentially, we pass the VLM three images: original $\boldsymbol{x}$, reconstruction A denoted $\boldsymbol{\hat x_0^A}$, and reconstruction B denoted $\boldsymbol{\hat x_0^B}$. We prompt the VLM to produce a numerical rating between -5 and 5 indicating whether reconstruction A or reconstruction B is closer to the original image. Negative numbers indicate A is superior. Prior to producing the numerical rating, the VLM is asked to provide detailed reasoning explaining the contents of each image and noting artifacts or inconsistencies in both reconstructions. The full text of the prompt is given in the supplementary material.

Since VLMs are prone to hallucination and to ignoring the contents of the provided images \cite{fu2025hidden}, we apply several mitigation strategies to improve the reliability of the reward signal. We do the following:

\begin{enumerate}
    \item For a given random seed $i$, we rate each pair of reconstructed images in two orders, reversing the order the second time, so the VLM produces rewards 
    \[ r_{B,0}^i = -r_{A,0}^i = \mathrm{VLM}(\boldsymbol{x}, \boldsymbol{\hat x_0^A}, \boldsymbol{\hat x_0^B}, i)\]
    and 
    \[ 
    r_{A,1}^i = -r_{B,1}^i = \mathrm{VLM}(\boldsymbol{x}, \boldsymbol{\hat x_0^B}, \boldsymbol{\hat x_0^A}, i).
    \] 
    The final ratings per seed are then given by 
    \[ r_A^i = \mathrm{sign}(r_{A,0}^i + r_{A,1}^i), ~~r_B^i = \mathrm{sign}(r_{B,0}^i + r_{B,1}^i) .\]
    \item Ensemble the reward over $n$ random seeds of the VLM, so that 
    \[
    r_A = \sum_{i=1}^n r_A^i,~~r_B = \sum_{i=1}^n r_B^i
    \]
    where $r_A^i$ is the rating assigned to image A for seed $i$, and similarly for $r_B^i$ respectively.
    \item Ensemble the VLM reward with LPIPS~\cite{zhang2018unreasonable}, a traditional perceptual metric. We require LPIPS and the VLM to produce a unanimous judgment in order to use a preference pair for training. If they disagree, the example is discarded.
\end{enumerate}

These modifications help to reduce the noise in the VLM reward computation and provide a more consistent training signal. Importantly, ensembling with LPIPS provides superior performance to using LPIPS or the VLM reward alone, as we show later in experiments. 

\begin{figure*}
    \includegraphics[width=\linewidth]{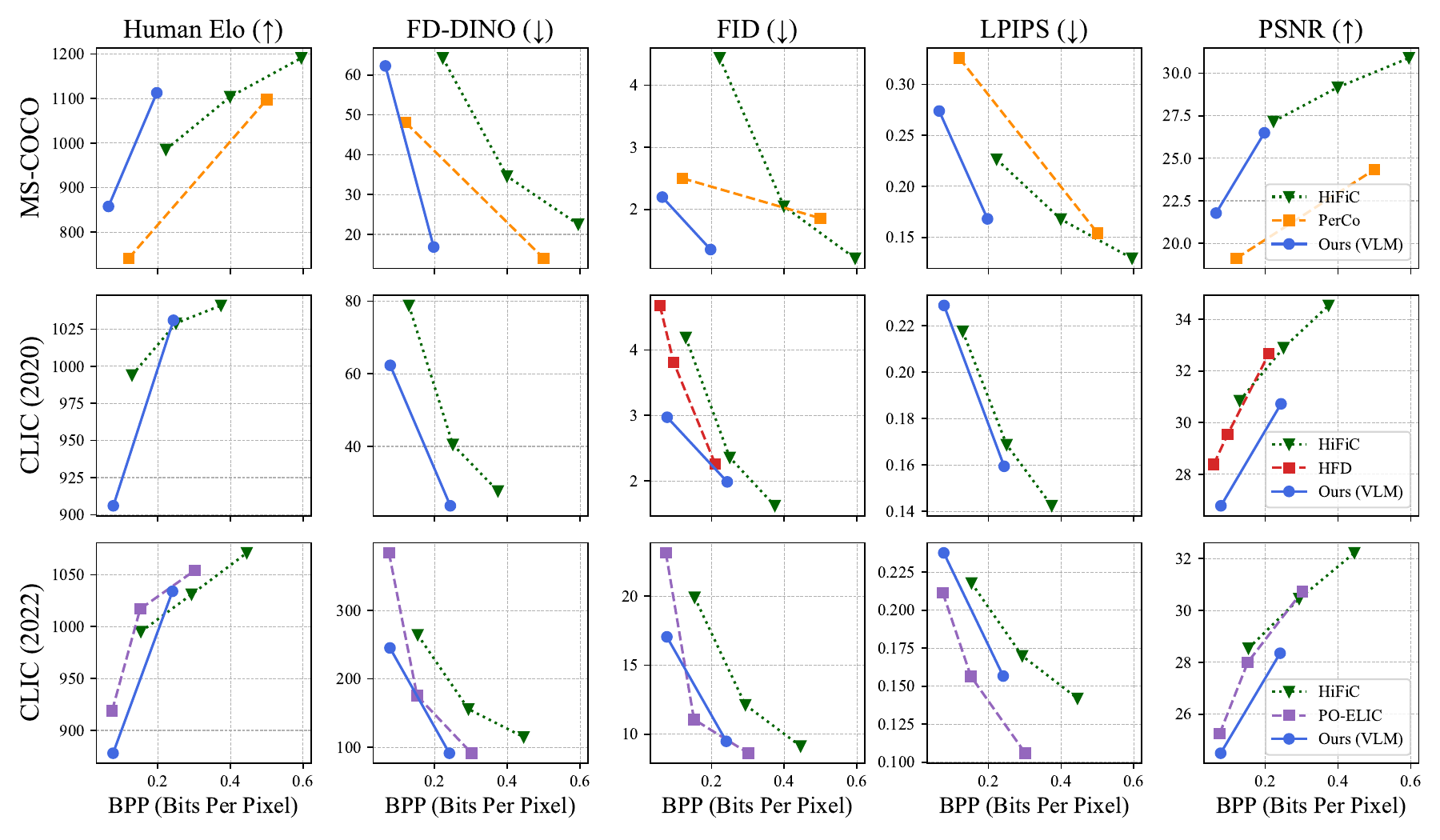}
     \vspace{-9mm}
     \caption{\textbf{Quantitative Evaluation on Image Compression Datasets.} Overall, \modelname~achieves competitive or state-of-the-art performance. \modelname~performs particularly well on perceptual metrics and particularly well on MS-COCO, which contains a high percentage of images with human-relevant characteristics such as text and faces.}
    \label{fig:main_results}
    \vspace{-4mm}
\end{figure*}

\section{Experiments}
\label{sec:experiments}
In the following experiments, we provide some technical details on our model training and setup. Then we present our main results comparing against state-of-the-art compression baselines on standard benchmark datasets. We contextualize and interpret the results as they are presented. Finally, we provide some additional ablation studies and analysis experiments.

\noindent\paragraph{Our setup.} We train \modelname~models at two BPP values: $0.07$, and $0.21$. We pretrain all \modelname~models for $1,000,000$ steps with Adam~\cite{kingma2014adam} learning rate $10^{-4}$. DPO posttraining runs for $8,000$ steps with learning rate $5\times 10^{-7}$. The batch size is $256$ in both stages. Training was completed on 256 TPUv4. All models were trained in JAX \cite{jax2018github} with bfloat16 precision.  We leverage Diffusion DPO in an online fashion, sampling a preference buffer of approximately $2,560$ examples every 250 steps (some examples may be discarded for the ensemble reward). We found online training with updated buffers provided superior results compared with synthesizing an offline preference dataset. We train using VLM rewards in an asynchronous fashion, simultaneously querying the VLM over the network using an updated sample buffer and performing DPO training on a slightly out-of-date sample buffer, so the latency of the VLM computation can be overlapped with DPO training.

Our models were trained at $256 \times 256$ resolution on ImageNet \cite{deng2009imagenet}. To accommodate variable resolutions, we use a tiled inference procedure similar to prior work \cite{hoogeboom2023hfd}, the details of which can be found in the supplementary material. We use a shifted schedule during inference and implement classifier-free guidance by dropping out the discrete latent code $10\%$ of the time, following prior work \cite{sargent2025flowmodemodeseekingdiffusion}.

\subsection{Main results.}
First, we will present the baselines, datasets, and metrics. Then we will analyze and explain the main results. 

\vspace{-5mm}
\noindent\paragraph{Baselines.} We compare against HiFiC~\cite{mentzer2020high}, PerCo~\cite{careil2023perco}, HFD~\cite{hoogeboom2023hfd}, and PO-ELIC~\cite{he2022poelic}. Since not all baselines have released code or reconstructions, and not all baselines can operate at all resolutions, we provide numbers only where possible for each baseline. We provide metrics for HiFiC for all datasets and metrics since the code is public. PerCo does not have released code, and after multiple attempts to correspond with the authors, we have instead chosen to rely on an open source replication \cite{Korber2024OpenPerco} which matches the performance on most perceptual metrics. A limitation of PerCo is that it cannot handle high resolution data such as CLIC 2020 and CLIC 2022. HFD has not released code or reconstructions, but we have taken their CLIC 2020 numbers from the paper after exactly replicating their evaluation pipeline on CLIC 2020. PO-ELIC has not released code, but has released reconstructions on CLIC 2022, so we directly compare against the released reconstructions.

\vspace{-4mm}
\noindent\paragraph{Datasets.} MS-COCO~\cite{lin2014microsoft} is a standard image compression benchmark which consists of a selection of $30,000$ images from the MS-COCO 2014 validation set. We crop these images to $256 \times 256$ using the DALL-E crop protocol following common practice \cite{hoogeboom2023hfd, mentzer2020high}. We compare against PerCo and HiFiC on MS-COCO.

The CLIC 2020 \cite{CLIC2020Dataset} train set is a standard image compression benchmark containing 428 high-resolution images of up to 4 megapixels. We compare against HiFiC and HFD on CLIC 2020. 

The CLIC 2022~\cite{CLIC2022Dataset} test set is a standard compression benchmark of 30 high-resolution images of up to 4 megapixels. We compare against PO-ELIC and HiFiC on CLIC 2022.
\vspace{-3mm}

\begin{table*}
\centering
\small{
\begin{tabular}{lccccc}
\toprule
 MS-COCO & Human Elo $\uparrow$ &  FD-DINO $\downarrow$& FID $\downarrow$ & LPIPS $\downarrow$ & PSNR $\uparrow$\\
 \midrule
Ours (0.07bpp) & \textbf{858} & \textbf{62.25} & \textbf{2.20} & \textbf{0.274} & \textbf{21.78} \\
~~$-$ LPIPS post-training only   & 838 &  63.63  & 2.33 & \textbf{0.274} & 21.77 \\
\cmidrule(lr){1-6}
Ours (0.21bpp) & \textbf{1112} &  \textbf{16.83}  & 1.35 & \textbf{0.168} & 26.50 \\
~~$-$ LPIPS post-training only & 1103 & 16.96  & \textbf{1.30} & 0.169 & \textbf{26.54} \\
\bottomrule
\end{tabular}
}
\vspace{-2mm}
\caption{\textbf{Importance of VLM.} At multiple BPP (prior to entropy coding), post-training with the VLM + LPIPS objective provides gains over post-training the compression model with LPIPS alone.}
\label{tab:lpips_h2h}
\vspace{-4mm}
\end{table*}

\begin{table}
\centering
\small{
\begin{tabular}{lcc}
\toprule
 Accuracy & BAPPS-Val & Compressed Images \\
 \midrule
Human & 73.99 & 72.15 \\
\cmidrule(lr){1-3}
LPIPS & 69.56 & 92.32 \\
DreamSim & 68.13\rlap{\textsuperscript{\(\dagger\)}} & - \\
VLM & 69.44 & 83.80 \\
\bottomrule
\end{tabular}
}
\vspace{-1mm}
\caption{\textbf{Human 2AFC benchmarks.} Gemini 2.5-Flash can replicate human judgments on 2AFC datasets zero-shot. $^\dagger$\textit{Number taken from paper}. 
}

\label{tab:judgment_replication}
\vspace{-6mm}
\end{table}

\noindent\paragraph{Metrics.} 
For all datasets, we evaluate with standard image quality metrics, namely LPIPS \cite{zhang2018unreasonable} and PSNR. We additionally compute two distributional image quality metrics, FID \cite{heusel2017gans}, and FD-DINO \cite{stein2023exposing}. Finally, we compute human Elo via large-scale user-studies. 

For high-resolution datasets such as CLIC 2020 and CLIC 2022, we evaluate distributional metrics such as FID~\cite{heusel2017gans} and FD-DINO~\cite{stein2023exposing} on square $256 \times 256$ random crops, following prior work \cite{mentzer2020high, hoogeboom2023hfd}. To maintain a sufficient sample size, we use 100 random crops per image for CLIC 2022 and 10 random crops per image for MSCOCO.

For each dataset, we conduct large-scale user studies of the compressed images, in which users are asked to visually compare two compressed versions of the same image. We collect 1,812 pairwise ratings for CLIC 2020, 2,523 pairwise ratings for CLIC 2022, and 15,705 pairwise ratings for MS-COCO. Since Elo~\cite{Elo1978} is order-dependent but our ratings per dataset are collected in a single parallelized job, we report the Elo averaged over 10,000 random re-shufflings of the rating order. For CLIC 2020 and CLIC 2022, due to the very high resolution, raters are shown random crops rather than full images. More details on the rating process are available in the supplementary material. 

We emphasize that for the evaluation of perceptually oriented visual compression, prior work has shown perceptually oriented distortion metrics and simple distortion metrics like PSNR are at odds with each other given a fixed rate \cite{blau2019rethinking}. Therefore, it is theoretically challenging to achieve improvements on all metrics simultaneously, so metrics correlated with human perception should be given greater weight in our analysis, even if they come at the expense of less correlated metrics. In general, we consider Elo to be the gold standard, since we directly compute it from thousands of human ratings of the test images. Prior work has also noted that FD-DINO is more predictive of human judgments than FID \cite{stein2023exposing}. Finally, the poor correlation of PSNR with human assessment of perceptual quality at a given bitrate is well-known \cite{1292216}, and we regard it as the least important metric in our analysis.

\noindent\paragraph{Results.}

Overall, \modelname~achieves very strong performance, as shown in Figure~\ref{fig:main_results}. It generally achieves stronger performance than HiFiC, PerCo, and HFD. It achieves particularly strong performance on MS-COCO, which contains a high percentage of images with human-relevant features such as text and faces. \modelname~under-performs relative to PO-ELIC on CLIC 2022, but without released code, the performance of PO-ELIC on low-resolution data or other datasets is unclear, and it is important to note that CLIC 2022 contains only 30 images. 

Additionally, \modelname~tends to achieve relatively stronger metrics on human-correlated metrics (i.e., ELO, FD-DINO) compared with PSNR. For instance, \modelname~achieves superior perceptual metrics in general relative to HFD and HiFiC, but worse PSNR. We deem this appropriate and even desirable, since our goal is to maximize human perception of reconstruction quality and these metrics are at odds given a fixed BPP and assuming enough model capacity \cite{blau2019rethinking}.

\subsection{Analysis}
\vspace{-6mm}
\noindent\paragraph{Replicating human judgments.} To provide motivation for using a VLM to approximate a human perceptual judge, we show in Table \ref{tab:judgment_replication} that Gemini 2.5-Flash can be used to replicate human perceptual judgments on the BAPPS dataset \cite{zhang2018unreasonable} and on our own collected dataset of human preferences (``Compressed images''). Compressed Images contains 5401 tuples of compressed images from two random baselines or our method; each tuple is rated at least 3 times by human raters via a custom interface. Inter-rater agreement (the ``Human'' row) is measured by holding out one human judgment and measuring its agreement with the average of the remaining human judgments for that tuple. It is somewhat surprising that both LPIPS and the VLM exceed the performance of a single human on our compressed images, but this is attributable to the compressed images being highly similar to the original image (much more similar than distorted BAPPS images, on average), meaning human judgments are noisier on average. 
\vspace{-4mm}

\noindent\paragraph{Importance of the VLM.} Since our final VLM reward is ensembled with LPIPS, it is important to verify that the VLM provides gains over post-training with a binary LPIPS-determined reward alone. We provide this comparison in Table \ref{tab:lpips_h2h}. Adding in the VLM reward provides a noticeable performance boost. At lower bitrates, the VLM provides a stronger reward signal since VLM judgments are less noisy when the images are more different. 

\noindent\paragraph{Mitigating noise.} The noisiness of the VLM reward on highly similar images is a serious issue. Figure \ref{fig:failure} shows a characteristic failure mode where, for highly similar reconstructions, the VLM fails to be self-consistent when the order of reconstructed images is reversed.

\begin{figure}
    \centering
    \includegraphics[width=\linewidth]{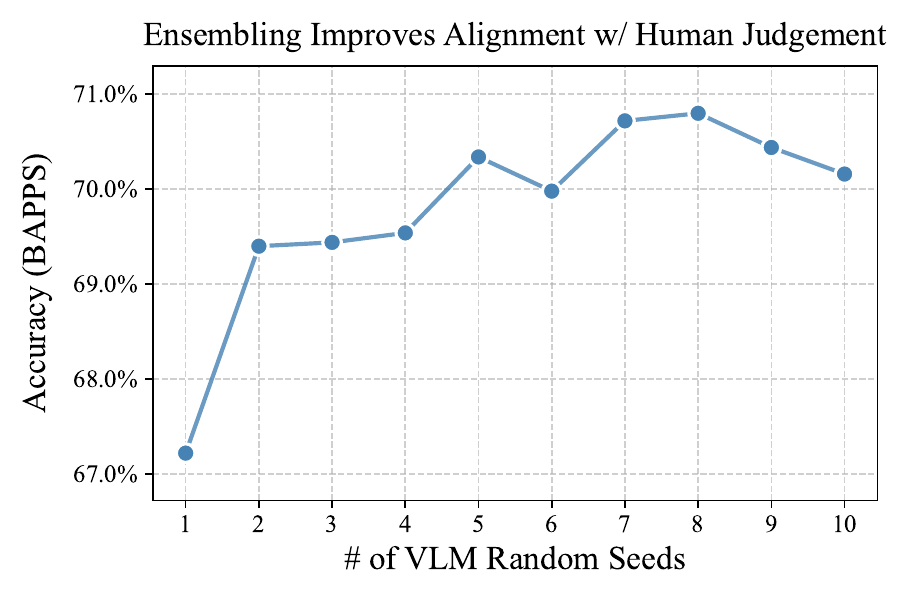}
    \vspace{-7mm}
    \caption{\textbf{Scaling self-ensembling.} The VLM becomes more predictive of human judgment on BAPPS \cite{kettunen2019elpips} as test-time compute (number of VLM seeds) is scaled.}
    \label{fig:ensemble}
\end{figure}
Indeed, reducing the noise in the VLM judgment is critical, especially since DPO struggles with noisy judgments \cite{wu2024drdpo}. Thankfully, the noisiness of the VLM reward can be mitigated in several ways. Most helpful is self-ensembling, \ie, computing the reward as the majority vote of multiple captioning requests to the VLM, ensembled with itself over multiple random seeds. The importance of self-ensembling is reflected in Figure \ref{fig:ensemble}, where we observe that performance on human judgment data increases as the number of VLM random seeds is increased, though eventually saturating. For our main experiments, we use $n=3$ seeds, though more seeds would improve performance at the cost of more VLM queries.

\begin{table}
\scriptsize{
\begin{tabular}{lcccc}
\toprule
 MS-COCO & FD-DINO $\downarrow$ & FID $\downarrow$ & LPIPS $\downarrow$ & PSNR $\uparrow$\\
 \midrule
Ours (\modelname) & 67.83 & 2.31  & \textbf{0.278} & \textbf{21.68} \\
$-$ No ensemble w/ LPIPS & \textbf{67.68} & \textbf{2.10} & 0.280 & 21.29\\
$-$ No post-training & 82.31 & 2.40 & 0.300 & 21.27 \\
$-$ No self-ensembling & 68.36 & 2.15 & 0.280 & 21.53 \\
\bottomrule
\end{tabular}
}
\vspace{-2mm}
\caption{\textbf{Ablation study.} Various components of reward design are necessary for best performance.
}
\label{tab:ablation}
\vspace{-3mm}
\end{table}
\noindent\paragraph{Ablation study.} In order to verify various other components of the system, we provide a large-scale ablation study in Table \ref{tab:ablation}. This study is conducted on MS-COCO. Interpreting the table, we see that only using the VLM to rank reconstructed images yields comparable performance to ensembling LPIPS and the VLM together (``No ensembling with LPIPS''), with the main difference being worse pixel-aligned metrics, namely PSNR and LPIPS, though distributional metrics actually improve. Failing to post-train the model with DPO yields poorer performance on every single metric. Not ensembling the VLM reward with itself (``No self-ensembling'') yields poorer performance on the majority of metrics, as the reward becomes noisier.

\begin{figure}
    \centering
    \includegraphics[width=\linewidth]{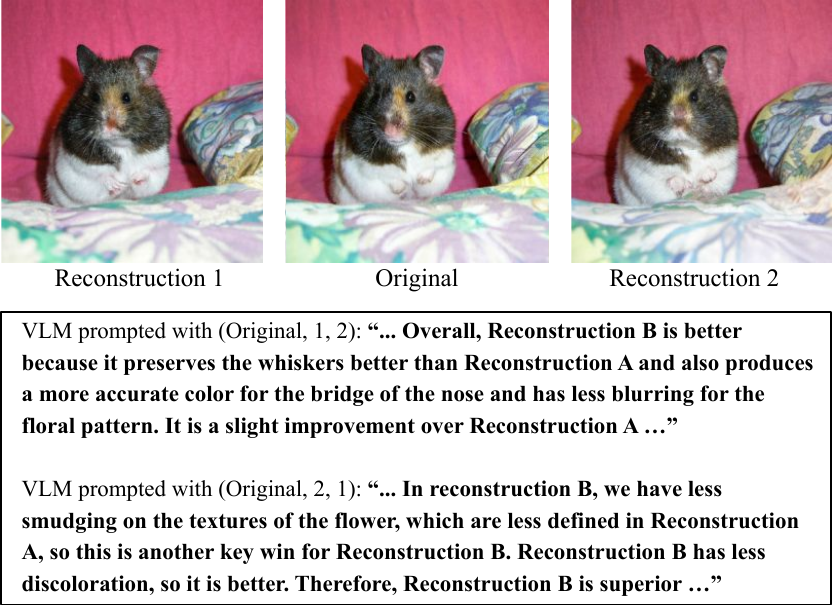}
    \caption{\textbf{Failure modes.} VLMs can hallucinate an incorrect ranking when the images are highly similar, such as in this case when the VLM fails to be self-consistent when the order of reconstructed images is reversed.}
    \label{fig:failure}
\vspace{-5mm}
\end{figure}

\paragraph{Limitations.}
The diffusion decoder adds additional latency compared with GAN-based compression methods, though this limitation is not unique to our method and is shared by other diffusion-based approaches \cite{careil2023perco, hoogeboom2023hfd, yang2023mandt, theis2022diffc}. Additionally, VLM-based rewards are more expensive to compute than evaluations of a small perceptual network.

\section{Conclusion}
In this paper, we showed that off-the-shelf VLMs have learned a visual prior that is highly correlated with human perception. When prompted to reason about the differences between images, we showed that VLMs can replicate human similarity judgments. Motivated by this, we designed a diffusion-based compression system, \modelname, designed to be trained with VLM preferences. We then post-trained the system with VLM preferences, and achieved competitive or state-of-the-art performance on human-aligned image compression depending on the dataset. 

The quality of \modelname~is dependent on the accuracy of the VLM used as a perceptual judge, and as VLMs are improved through considerable research and investment, image compression techniques such as \modelname~may benefit from their stronger zero-shot perceptual priors and achieve further improvements in human-aligned compression performance.

\noindent\paragraph{Acknowledgments.} We thank Ben Poole, David Minnen, and Dina Bashkirova for helpful discussions. 
{
    \small
    \bibliographystyle{ieeenat_fullname}
    \bibliography{main}
}

\clearpage
\setcounter{page}{1}
\maketitlesupplementary
\appendix

This supplementary material contains details regarding models, training, inference, user studies, and the VLM prompt. For more visualizations, please consult the attached supplementary webpage. Thank you!

\section{Models and training}

Our model is fully transformer-based, consisting of a transformer encoder and decoder and an autoregressive transformer entropy coder. We provide the relevant hyperparameters for the encoder, decoder, and autoregressive model below. The encoder and decoder are identical to FlowMo \cite{sargent2025flowmodemodeseekingdiffusion}, with the exception that we use FSQ quantization \cite{Mentzer2023FSQ}. The entropy coder is based on NanoGPT \cite{nanogpt}. The sequence length for the encoder and decoder is computed as the sum of the number of image tokens and latent tokens.

\begin{table}[h]
    \centering
    \begin{tabular}{cccc}
        \toprule
        Low BPP & Encoder & Decoder & Entropy coder\\
         \midrule
        Hidden dim. & $768$ & $1152$ & $768$ \\
        Number of layers & $8$ & $16$ & $16$\\
        Patch size & $4$ & $4$ & - \\
        1D Token size & $6$ & $6$ & - \\
        FSQ levels & $8$ & $8$ & - \\
        Sequence length & $4352$ & $4352$ & $384$ \\ \bottomrule
    \end{tabular}
    \caption{Model hyperparameters for low BPP configuration. Total parameter count is 1.01B}
    \label{tab:placeholder}
\end{table}

\begin{table}[h]
    \centering
    \begin{tabular}{cccc} 
    \toprule
    High BPP & Encoder & Decoder & Entropy coder\\
         \midrule
        Hidden dim. & $768$ & $1152$ & $768$ \\
        Number of layers & $8$ & $16$ & $16$\\
        Patch size & $4$ & $4$ & - \\
        1D Token size & $18$ & $18$ & - \\
        FSQ levels & $8$ & $8$ & - \\
        Sequence length & $4352$ & $4352$ & $1152$ \\ \bottomrule
    \end{tabular}
    \caption{Model hyperparameters for high BPP configuration. Total parameter count is 1.01B}
    \label{tab:placeholder}
\end{table}

Training proceeds in three stages. All trainings are done on 256 TPUv4 on ImageNet \cite{deng2009imagenet} with resolution 256 and with batch size 256. We use a random horizontal flip and center cropping. We provide details on each stage below:

\begin{enumerate}
    \item Pretraining is done for $1,000,000$ steps with batch size 256. We use the Adam optimizer with learning rate $10^{-4}$ \cite{kingma2014adam} and no weight decay or dropout.
    \item Post-training is done via Diffusion DPO with learning rate $5 \times 10^{-7}$ using the desired reward (LPIPS, VLM, VLM ensembled with LPIPS, etc.). The DPO sample buffers are recomputed every $250$ steps and contain $2,560$ samples. 
    \item The entropy coder is trained for $200,000$ steps with learning rate $10^{-4}$, dropout with $p=0.1$ and weight decay $0.025$ with AdamW. The encoder and decoder are frozen during this period.
\end{enumerate}

\section{VLM prompt}
The full prompt for Gemini 2.5 Flash is shown below: \newline

\textit{``In this task, you’ll be asked to compare an original image with two AI reconstructions of that image, Reconstruction A followed by Reconstruction B. You’ll see triplets of images like (original, A, B). You’ll give a relative rating of the two images. This is between -5 and 5, inclusive. Higher scores mean that Reconstruction B is relatively better. So if you give a score of -2, you think A is kind of better, and if you give a score of +5, you think B is obviously significantly better. Be sparing with the higher magnitude scores - you should expect that most triplets you see won't have an obviously better reconstruction. In your ratings, make sure to prioritize differences that are semantically important to a human observer. If a distortion changes the meaning of an image to a human observer, then it's more significant than if a distortion changes the texture of an image. Your response should conclude with ``RATING: X", where X is your rating, i.e. -2 or 3. Now here is the image triplet that you need to rate. Make sure to provide a lot more reasoning and make sure to carefully look at each of the three images and provide meticulous visual justifications based on evidence from each image! Remember, negative scores mean that A is better, and positive scores mean that B is better.''}
\newline
\newline
Once the response is produced, the reward is computed as described in the method section, given the final numerical rating.

\begin{figure*}[]
    \centering
    \begin{tabular}{cc}
         \includegraphics[width=0.6\linewidth,height=0.4\linewidth]{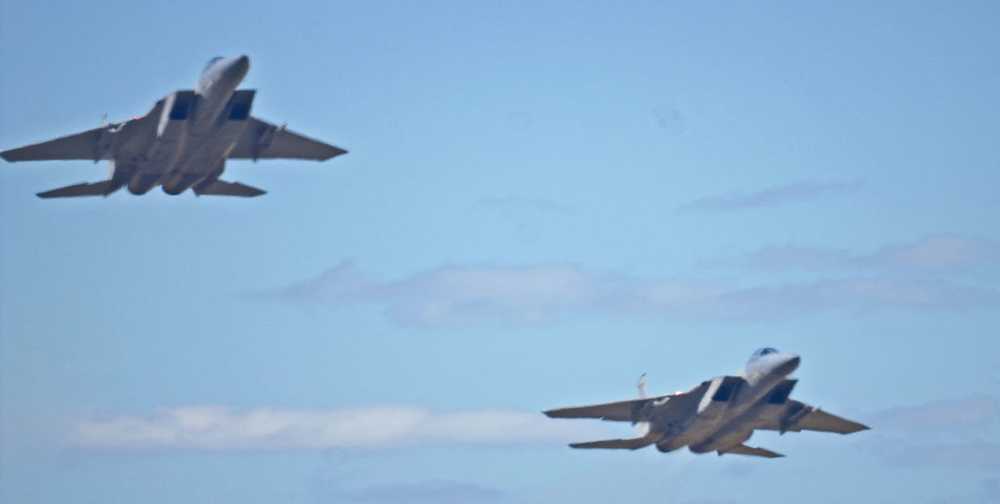} \\
         \includegraphics[width=0.6\linewidth,height=0.4\linewidth]{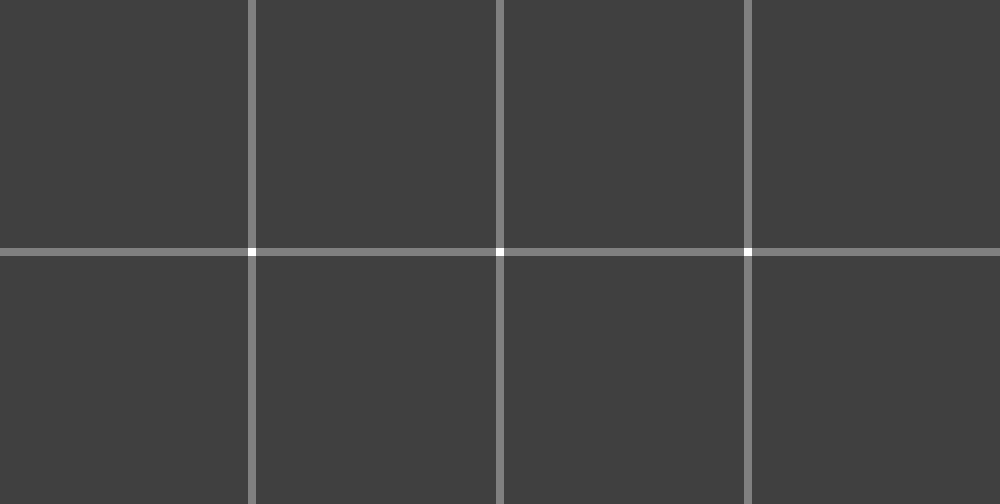} \\
         \includegraphics[width=0.6\linewidth]{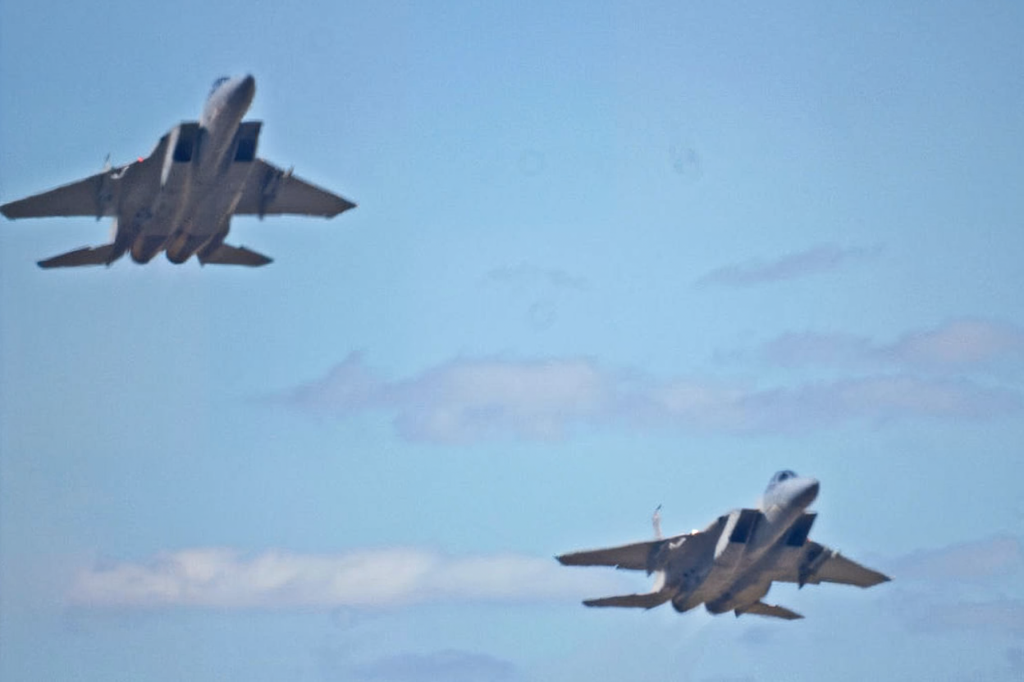} \\
    \end{tabular}
    \caption{\textbf{Tiled inference for arbitrary resolutions.} From top to bottom: Original image, tiling strategy, reconstructed image. The margin size (we use 8 pixels in this work) must be large enough to communicate information between patches during diffusion to avoid unsightly border artifacts, but not so large as to waste BPP.}
    \label{fig:tiling}
\end{figure*}

\section{Inference}
Since our model is only trained on $256\times 256$ images, a zero-shot procedure is needed to support inference at higher resolutions at test time. A visualization of the tiled inference strategy we use to support any-resolution compression is shown in Figure \ref{fig:tiling}. Where the image dimensions are $h,w$, $r$ is the native model resolution and $p$ is the margin size, we compute the smallest $k_h, k_w$ such that $r + k_h(r - p) \geq h$ and $r + k_w(r - p) \geq w$. The image is then resized to $r + k_h(r - p), r + k_w(r - p)$ and broken into overlapping tiles, which are separately encoded. The image tiles are then diffused jointly following MultiDiffusion \cite{bar2023multidiffusion}.

\section{User studies}

For policy-related reasons, we cannot share visuals of the exact user interface. Users are presented with three images: Image A, Original Image, and Image B (in that order). The user is asked to select an answer from 5 options:
\newline 

\textit{
Which image is more similar to the original image?}

\begin{itemize}
    \item \textit{Image A is much more similar}
    \item \textit{Image A is slightly more similar}
    \item \textit{About the same}
    \item \textit{Image B is slightly more similar}
    \item \textit{Image B is much more similar}\newline
\end{itemize}

Since users may have varying display sizes, images are resized to 480 pixels in width. We show random crops for CLIC 2020 and CLIC 2022 due to the very high resolution.

\section{Additional capabilities}
\begin{figure*}
    \centering
    \includegraphics[width=\linewidth]{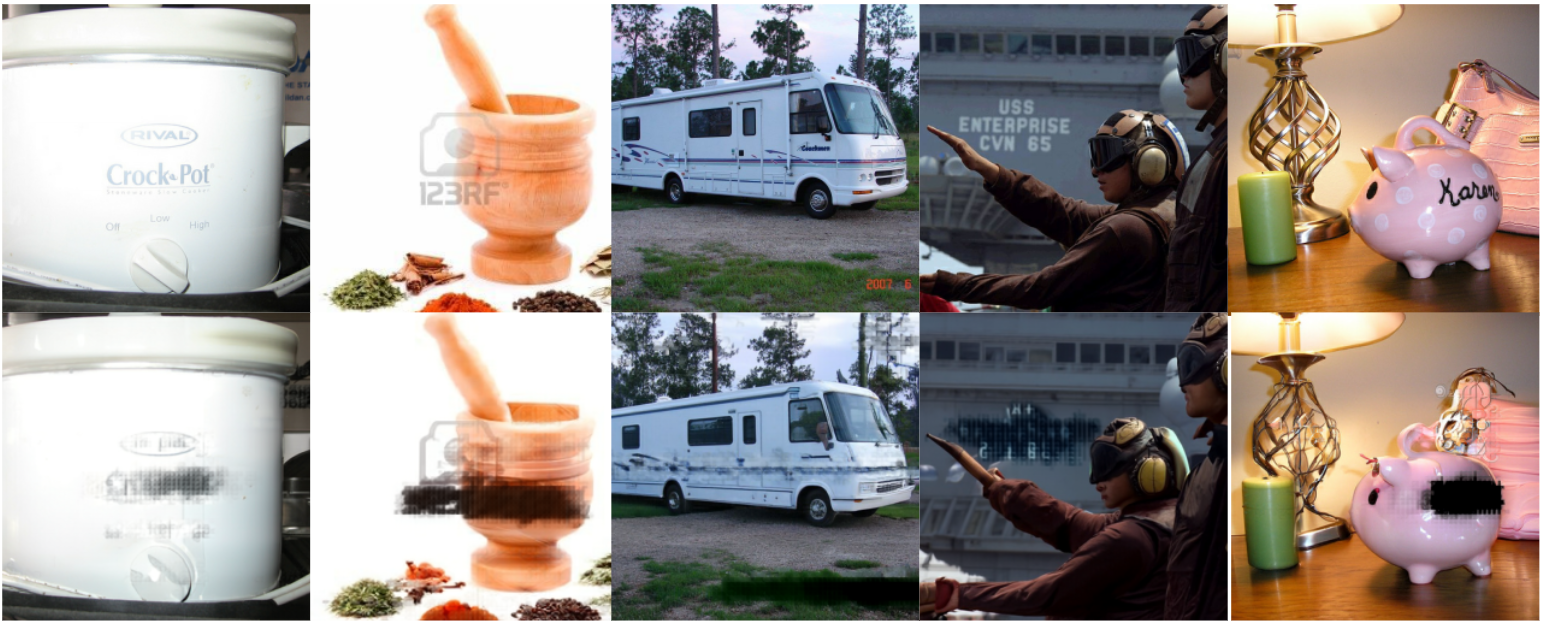}
    \caption{\textbf{Censoring readable text.} A failure case of an edit-distance based reward on readable text determined by the VLM causes the model to degenerate to censoring all readable text in the images.}
    \label{fig:censoring}
\end{figure*}

Since Diffusion DPO only requires binary preferences, we may leverage the VLM to rate reconstructed images according to criteria other than perceptual quality. For instance, in Figure \ref{fig:censoring}, we instruct the VLM to read out the readable text in the original and reconstructed image, and compute the reward as the edit distance between the text in the original and reconstructed image. In this case, the model degenerates to a local minimum where the image text is censored, so the readable text in the reconstruction has edit distance bounded by the length of the text in the original image. Alternative formulations of edit distance can lead to slightly improved performance on average for rendering readable text, though the difference is not generally noticeable.

\section{Disclaimer} Some images in this paper contain content which is sensitive to left-right orientation such as readable text. While some standard benchmark datasets such as MS-COCO are prepared with random flips, we may re-flip the reconstructions of both ours and the baselines for ease of visualization.

\section{Image attributions}


\noindent \textbf{Figure 1, Truck}
\begin{itemize}
    \item Sourced From: ImageNet
    \item Image ID: \texttt{n03594945\_15055}
    \item License: \href{https://image-net.org/accessagreement}{ImageNet Agreement}
\end{itemize}

\noindent \textbf{Figure 1, Street Sign}
\begin{itemize}
    \item Sourced From: COCO
    \item Image ID: \texttt{000000001779}
    \item Original Creator: \href{https://www.flickr.com/people/isfullofcrap/}{I am R.}
    \item License: \href{https://creativecommons.org/licenses/by/2.0/}{CC BY 2.0}
    \item URL: \url{https://www.flickr.com/photos/isfullofcrap/2368878795/}
\end{itemize}

\noindent \textbf{Figure 1, Bar}
\begin{itemize}
    \item Sourced From: COCO
    \item Image ID: \texttt{000000000723}
    \item Original Creator: \href{https://www.flickr.com/photos/hankzby/}{Henry Zbyszynski}
    \item License: \href{https://creativecommons.org/licenses/by/2.0/}{CC BY 2.0}
    \item URL: \url{https://www.flickr.com/photos/hankzby/7385695522/}
\end{itemize}

\noindent \textbf{Figure 2, Chicago Skyline}
\begin{itemize}
    \item Sourced From: CLIC 2022
    \item Image ID: \texttt{07113e38700d3f0dab7a9f34d4512 98a54de3cef3bc4e03945d5fead4f513ecd}
    \item License: \href{https://help.unsplash.com/en/collections/1463188-unsplash-license}{Unsplash}
\end{itemize}

\noindent \textbf{Figure 2, Street Sign}
\begin{itemize}
    \item Sourced From: COCO
    \item Image ID: \texttt{000000000250}
    \item Original Creator: Unknown
    \item License: \href{https://creativecommons.org/licenses/by-nc-sa/2.0/}{CC BY-NC-SA 2.0}
    \item URL: \url{http://images.cocodataset.org/train2014/COCO_train2014_000000000250.jpg}
\end{itemize}

\noindent \textbf{Figure 2, Women on Phone}
\begin{itemize}
    \item Sourced From: COCO 
    \item Image ID: \texttt{000000000536}
    \item Original Creator: Unknown
    \item License: \href{https://creativecommons.org/licenses/by-nc-sa/2.0/}{CC BY-NC-SA 2.0}
    \item URL: \url{http://images.cocodataset.org/val2014/COCO_val2014_000000000536.jpg}
\end{itemize}

\noindent \textbf{Figure 2, Soldiers}
\begin{itemize}
    \item Sourced From: COCO
    \item Image ID: \texttt{000000001149}
    \item Original Creator: \href{https://www.flickr.com/photos/picatinnyarsenal/}{Picatinny Arsenal}
    \item License: \href{https://creativecommons.org/licenses/by-nc/2.0/}{CC BY-NC 2.0}
    \item URL: \url{https://www.flickr.com/photos/picatinnyarsenal/5202323780/}
\end{itemize}

\noindent \textbf{Figure 3, Gazelles}
\begin{itemize}
    \item Sourced From: ImageNet
    \item Image ID: \texttt{n02423022\_31692}
    \item License: \href{https://image-net.org/accessagreement}{ImageNet Agreement}
\end{itemize}

\noindent \textbf{Figure 6, Mouse}
\begin{itemize}
    \item Sourced From: ImageNet
    \item License: \href{https://image-net.org/accessagreement}{ImageNet Agreement}
\end{itemize}

\noindent \textbf{Gallery, Bridge}
\begin{itemize}
    \item Sourced From: CLIC 2022
    \item Image ID: \texttt{732bf474788c19c0c1fae6dd7689d4c
    da2f4e0632a1c7725e970b69d44d08f3e}
    \item License: \href{https://help.unsplash.com/en/collections/1463188-unsplash-license}{Unsplash}
\end{itemize}

\noindent \textbf{Gallery, Tennis player}
\begin{itemize}
    \item Sourced From: COCO
    \item Image ID: 000000001815
    \item Original Creator: \href{https://www.flickr.com/photos/43555660@N00/}{Kate Tann}
    \item License: CC BY-SA 2.0
    \item URL: \url{https://www.flickr.com/photos/43555660@N00/6050671677/}
\end{itemize}

\noindent \textbf{Gallery, Woman}
\begin{itemize}
    \item Sourced From: COCO
    \item Image ID: 000000001569
    \item Original Creator: \href{https://www.flickr.com/photos/tomconger/}{Tom Conger}
    \item License: CC BY-NC 2.0
    \item \url{https://www.flickr.com/photos/tomconger/3578082722/}
\end{itemize}

\noindent \textbf{Gallery, Hunter}
\begin{itemize}
    \item Sourced From: COCO
    \item Image ID: 000000001948
    \item Original Creator: Unknown
    \item License: CC BY-NC-SA 2.0
    \item \url{http://images.cocodataset.org/val2014/COCO_val2014_000000001948.jpg}
\end{itemize}

\noindent \textbf{Gallery, Dinner}
\begin{itemize}
    \item Sourced From: CLIC 2020
    \item License: \href{https://help.unsplash.com/en/collections/1463188-unsplash-license}{Unsplash}
\end{itemize}

\noindent \textbf{Gallery, Lady in dress}
\begin{itemize}
    \item Sourced From: CLIC 2020
    \item License: \href{https://help.unsplash.com/en/collections/1463188-unsplash-license}{Unsplash}
\end{itemize}

\noindent \textbf{Gallery, Cat}
\begin{itemize}
    \item Sourced From: CLIC 2020
    \item License: \href{https://help.unsplash.com/en/collections/1463188-unsplash-license}{Unsplash}
\end{itemize}

\noindent \textbf{Gallery, Firehouse}
\begin{itemize}
    \item Sourced From: CLIC 2020
    \item License: \href{https://help.unsplash.com/en/collections/1463188-unsplash-license}{Unsplash}
\end{itemize}

\end{document}